\documentclass[conference]{IEEEtran}
\IEEEoverridecommandlockouts
\usepackage{cite}
\usepackage{amsmath,amssymb,amsfonts}
\usepackage{algorithmic}
\usepackage{graphicx}
\usepackage{textcomp}
\usepackage{framed,multirow,tabu}
\usepackage{xcolor}
\def\BibTeX{{\rm B\kern-.05em{\sc i\kern-.025em b}\kern-.08em
    T\kern-.1667em\lower.7ex\hbox{E}\kern-.125emX}}
\begin{document}
\title{A Graph-based approach to derive the geodesic distance on Statistical manifolds: Application to Multimedia Information Retrieval\\
}

\author{\IEEEauthorblockN{Zakariae Abbad}
\IEEEauthorblockA{\textit{LISAC} \\
\textit{FSDM, Sidi Mohamed Ben Abdellah University}\\
Fez, Morocco \\
zakariaeabbad@gmail.com}
\and
\IEEEauthorblockN{Ahmed Drissi El Maliani}
\IEEEauthorblockA{\textit{LRIT Rabat IT Center} \\
\textit{Faculty of sciences, Mohammed V University}\\
Rabat, Morocco \\
maliani.ahmed@gmail.com}
\and
\IEEEauthorblockN{Said Ouatik El Alaoui}
\IEEEauthorblockA{\textit{Laboratory of Engeneering Sciences} \\
\textit{National School of Applied Sciences, Ibn Tofail University}\\
Kenitra, Morocco \\
ouatikelalaoui.said@uit.ac.ma}
\and
\IEEEauthorblockN{Mohammed El Hassouni}
\IEEEauthorblockA{\textit{DESTEC, FLSHR, LRIT IT Center} \\
\textit{University of Mohammed V}\\
Rabat, Morocco \\
mohamed.elhassouni@gmail.com}
}
\IEEEoverridecommandlockouts \IEEEoverridecommandlockouts\IEEEpubid{\makebox[\columnwidth]{ 978-1-7281-9015-0/20/\$31.00~\copyright~2020~IEEE } \hspace{\columnsep}\makebox[\columnwidth]{ }}
\maketitle

\begin{abstract}
In this paper, we leverage the properties of non-Euclidean Geometry to define the Geodesic distance (GD) on the space of statistical manifolds. The Geodesic distance is a real and intuitive similarity measure that is a good alternative to the purely statistical and extensively used Kullback-Leibler divergence (KLD). Despite the effectiveness of the GD, a closed-form does not exist for many manifolds, since the geodesic equations are  hard to solve. This explains that the major studies have been content to use numerical approximations. Nevertheless, most of those do not take account of the manifold properties, which leads to a loss of information and thus to low performances. We propose an approximation of the Geodesic distance through a graph-based method. This latter permits to well represent the structure of  the statistical manifold, and respects its geometrical properties. 
 Our main aim is to compare the graph-based approximation to the state of the art approximations. Thus, the proposed approach is evaluated for two statistical manifolds, namely the Weibull manifold and the Gamma manifold, considering the Content-Based Texture Retrieval
 application on different databases.
\end{abstract}


\section{Introduction}

Statistical manifolds are geometric representations of smooth families of probability density functions. A class of probability density functions (Pdf) is characterized by a vector $\theta$ that forms a coordinate system of continuous parameters and has geometrical properties as a result of local compositions of the distributions. So naturally, a manifold (S) that governs these pdfs is a collection of points (representations of the pdfs) with a coordinate system. This latter makes the one-to-one mapping from S to $\mathbb{R}^{n}$ \cite{amari}\cite{rice}.\\
Information geometry is a branch of differential geometry that gives a geometrical perspective to the probabilistic theory and statistics. The research on the geometrical properties of probability distributions started with the work of Fisher \cite{fisher} and then was developed in the literature (\cite{amari}, \cite{divergencegeo} and \cite{rice}). After that, information geometry has been utilized to study the geometry of the statistical models as in \cite{riemannianpriors} where the geometry of the univariate normal model was used in image classification. In \cite{attgeo}, the Generalized Gaussian distribution geometry was studied. Also, The study of $G_0^1$ distributions geometry was applied to region discrimination in \cite{toresgeo}. The information geometry was also used to learn finite mixture models, such as the mixture of Gaussians \cite{yongeo} and the mixture of multivariate Gaussians \cite{dodgeo}.\\
Among all (curvature, affine connection ,...), the similarity measurement (SM) between probability distributions remains the most prominent example of the advantage of the geometrical understanding. It permits to derive a real distance, namely the geodesic distance (GD) on top of the statistical manifold. The GD is a natural distance that governs the statistical manifolds and is represented by the shortest path that exists between two points on the manifold. In this context, there have been numerous studies to investigate the GD among other metrics such as in \cite{geert1}, \cite{geert2} where Geert et al. proposed the Geodesic Distance  as a similarity measurement between the zero-mean Multivariate Generalized Gaussian distributions (MGGD) in the context of texture retrieval. In \cite{csd}, the Cauchy-Schwarz divergence was used to estimate the similarity between the Mixtures of Generalized Gaussian distributions in the texture retrieval context. \\It is to note that the Kullback-Leibler divergence (KLD) is the most popular in the context of texture and image retrieval. KLD was used in numerous papers like in \cite{gammakld} where it was combined with the Gamma distribution. In \cite{wblkld}, the KLD was used in combination with the Weibull distribution in the same context of texture retrieval. However, the KLD is not considered a real distance on the probability space, because it does not satisfy the triangular inequality and it is not symmetric.\\
In this work, we use the GD as a similarity measurement insted of the KLD, in order to integrate a geometric thinking in the context of image retrieval and more generally in the context of Multimedia Information retrieval.\\We will compute the GD on the Gamma and the Weibull manifolds. Both manifolds have been  extensively studied, such as in \cite{dodgamma} 
,\cite{rice}, \cite{arwini} for the Gamma manifold, and
\cite{arwini2},
\cite{wblmanifapp1} for the Weibull manifold. 
\\Due to the cumbersomeness of the geodesic equations, there is no closed-form but approximations of the GD in both cases.  We propose a new approximation , namely the graph-based approximation to compute the GD, in order to provides more precision to the retrieval process. Until the finish date of this work, no endeavor to do so was made. Experiments on two well-known texture datasets show that the GD through the graph-based approximation achieves an obvious higher performance compared with the classic Kullback-Leibler divergence.\\
The structure of this paper is as follows: In the next section, we provide the geometrical properties of the  Gamma \& Weibull manifolds. In section 3, we provide the graph-based approach that we used to approximate the geodesic distance. In section 4, we present the experimental results of the texture retrieval using the GD and the KLD, before concluding in section 5.

\section{The Geometry of statistical manifolds}

\subsection{The Gamma manifold}
The Gamma manifold is defined by the parametric family of the probability density function using the scale-shape parametrization:\begin{equation} f(x; \alpha, \beta) = \frac{ x^{\beta - 1}}{\alpha^{\beta} \gamma(\beta)} e^{-(\frac{x}{\alpha})}. \end{equation}
where  $\gamma(.)$ is the standard Gamma function.
\\The Gamma parameters ($\alpha$ and $\beta$) can be estimated by the maximum likelihood estimation (MLE), by solving this equation: \begin{equation}  \hat{\theta} = \arg\underset{\theta}\max \log \prod_{i=1}^{n} f(x_{i}; \alpha, \beta) .\end{equation}
Which leads to the following system of equations: 
\begin{equation} \hat{\alpha} = \frac{1}{n \hat{\beta}} \sum_{i=1}^{n} x_{i},\end{equation}

\begin{equation}
\log(\hat\beta) - \frac{\gamma\prime(\hat{\beta})}{\gamma(\hat{\beta})} =  \log(\frac{1}{n} \sum_{i=1}^{n} x_{i}) - \frac{1}{n}\sum_{i=1}^{n} \log(x_{i}).
\end{equation}
The mean and the variance of the Gamma distribution, are respectively given by:

\begin{equation}
E(X) = \alpha\beta,
\end{equation}

\begin{equation}
Var(X) = \beta \alpha^{2}.
\end{equation}
In information geometry \cite{amari},
a statistical model  $\{p_{\theta}; \theta \in \Theta\}$ , where $\Theta \subset \mathbb{R}^{r}$, can be provided with a Riemannian geometry, that is determined by the Fisher information matrix \cite{ovidiu}:
\begin{equation} g_{ij}(\theta) = E\left\{  \dfrac{\partial \ln p(X|\theta)}{\partial \theta^{i}}\dfrac{\partial \ln p(X|\theta)}{\partial \theta^{j}} \mid \theta \right\} .\end{equation} with (i, j = 1, 2, ..., r).
This is calculated in the case of the Gamma manifold by the following matrix \cite{arwini}:

\[g_{ij}(\theta) = \begin{pmatrix}
\frac{\beta}{\alpha^{2}} & 0 \\
0 & \dfrac{\partial^{2} \ln(\Gamma)}{\partial\beta}-\frac{1}{\beta}\\
\end{pmatrix}.\]
For each $\alpha \in \mathbb{R}$, the $\alpha$-connection is the torsion-free affine with the components:

\begin{equation}
\Gamma_{ij,k}^{(\alpha)} = \frac{1-\alpha}{2} \partial_{i}\partial_{j}\partial_{k}\varphi(\theta).
\end{equation}

Where $\varphi(\theta) = \log(\Gamma(\beta))-\beta\log(\alpha)$ is the corresponding potential function.

So in the case of Gamma manifold it becomes:

\begin{equation}\begin{split}
\Gamma_{11,1}^{(\alpha)} = -\dfrac{(1-\alpha)\beta}{\alpha^{3}},\\
\Gamma_{12,1}^{(\alpha)} = \Gamma_{12,2}^{(\alpha)} = \dfrac{(1-\alpha)}{2\alpha^{2}},\\
\Gamma_{22,2}^{(\alpha)} = \dfrac{(1-\alpha)\psi^{''}(\beta)}{2}.
\end{split}\end{equation}

\subsection{The Weibull manifold}

The Weibull manifold is also defined by the parametric family of the probability density function using the scale-shape parametrization: \begin{equation} f(x; \lambda, \mu) = \frac{\mu}{\lambda} (\frac{x}{\lambda})^{\mu-1} e^{-(\frac{x}{\lambda})^{\mu}} .\end{equation}
 where  $\gamma(.)$ is the standard Gamma function.
\\Using the MLE algorithm, parameters $\lambda$ and $\mu$ are estimated by solving the following system of equations: \begin{equation} \hat{\lambda}^{\mu} = \frac{1}{n} \sum_{i=1}^{n} {x_{i}}^{\mu} ,\end{equation}

\begin{equation}
\hat{\mu}^{-1} = \frac{\sum_{i=1}^{n} {x_{i}}^{\mu} \ln (x_{i})}{\sum_{i=1}^{n} {x_{i}}^{\mu}} - \frac{1}{n} \sum_{i=1}^{n} \ln(x_{i}).
\end{equation}
The mean and the variance of the Weibull distribution, are respectively given by:

\begin{equation}
E(X) = \lambda\gamma(1+\frac{1}{\mu}),
\end{equation}

\begin{equation}
Var(X) = \lambda^{2}\left[ \Gamma\left( 1+\frac{2}{\mu}\right)  - \left( \Gamma\left( 1+\frac{1}{\mu}\right) ^{2}\right)  \right]. 
\end{equation}

The Fisher information matrix in this case is defined by:

\[g_{ij}(\theta) = \begin{pmatrix}
\frac{\mu^{2}}{\lambda^{2}} & \dfrac{\xi-1}{\lambda} \\
\dfrac{\xi-1}{\lambda} & \dfrac{\xi^{2}-2\xi+\frac{\pi^{2}}{6}+1}{\mu^{2}}\\
\end{pmatrix}.\]
Where $\xi$ is the Euler constant.
\\The $\alpha$-connection exists, but it has long analytical expression, so we just mention that the Weibull manifold has a constant Christoffel symbols \cite{arwini2}: 

\begin{equation}\begin{split}
\Gamma_{11}^{1} = \dfrac{6(\xi\mu-\mu-\frac{\pi^{2}}{6})}{\pi^{2}\lambda}, \Gamma_{11}^{2} = -\dfrac{\mu^{3}}{\pi^{2}\lambda^{2}},\\ \Gamma_{21}^{1} = \Gamma_{12}^{1} = \dfrac{6(\xi^{2}-2\xi+\frac{\pi^{2}}{6}+1)}{\pi^{2}\mu},\\ \Gamma_{21}^{2} = \Gamma_{12}^{2} = \dfrac{6\mu(1-\xi)}{\pi^{2}\lambda},\\
\Gamma_{22}^{1} = -\dfrac{6\lambda(1-\xi)(\xi^{2}-2\xi+\frac{\pi^{2}}{6}+1)}{\pi^{2}\mu^{3}},\\
\Gamma_{22}^{2} = -\dfrac{6(\xi^{2}-2\xi+\frac{\pi^{2}}{6}+1)}{\pi^{2}\mu}.
\end{split}\end{equation}

\section{Approximation of the geodesic distance using the graph-based approach}
\subsection{The geodesic distance}
In \cite{rao}, Rao proposed the Rao-Geodesic distance for computing similarity between distributions of a parametric family, all of whose members satisfy certain conditions \cite{atkinson}. The metric is based on a Riemannian geometry, and is described in terms  of the information matrix elements  the family. In fact, by considering that $g_{ij}(\theta)$ is strictly positive, for each $\theta \in \Theta $, a Riemannian metric on $\Theta$ is defined by: \begin{equation} ds^{2}(\theta) = \sum_{i, j=1}^{r} g_{ij}(\theta) d\theta^{i} d\theta^{j}. \end{equation} Once this metric is introduced, given two probability measures $P_{\theta_{1}}$ and $P_{\theta_{2}}$ which belong to the statistical manifold, the geodesic distance between $P_{\theta_{1}}$ and $P_{\theta_{2}}$ is defined as the Riemannian distance between $\delta(\theta_{1}, \theta_{2}) \in \Theta$, and is given by: 
\begin{equation}\delta(\theta_{1}, \theta_{2}) =  \Bigg\vert \int_{t_{1}}^{t_{2}} {\left[  \sum_{i, j=1}^{r} g_{ij}(\theta) \frac{{d\theta}^{i}}{dt} \frac{{d\theta}^{j}}{dt}\right]^{\frac{1}{2}} dt } \Bigg\vert \end{equation}
Particularly, among the curves between $\theta_{1}$ and $\theta_{1}$, we are interested by the one that represents the minimum distance between these two points. It is called the Geodesic, and it is given as a solution to differential equations, called the geodesic equations : \begin{equation}\ddot{\theta^{k}}(t) + \sum_{i, j}^{} {\Gamma^{k}_{ij}} [\theta(t)] \dot{\theta^{i}}(t) \dot{\theta^{j}}(t) = 0 .
\end{equation} where the ${\Gamma^{k}_{\mu\upsilon}}$ are the Christoffel symbols of the second kind, defined by:
\begin{equation}{\Gamma^{k}_{\mu\upsilon}} = \frac{1}{2} \sum_{\rho}^{} g^{k\rho} (\dfrac{\partial g_{\upsilon\rho}}{\partial \theta^{\mu}} + \dfrac{\partial g_{\mu\rho}}{\partial \theta^{\upsilon}} - \dfrac{\partial g_{\mu\upsilon}}{\partial \theta^{\rho}}) .\end{equation} and $g^{\mu\upsilon}$ denotes the components of the inverse metric.

\subsection{The graph-based approach}

In this work, the statistical manifolds are viewed as weighted graphe. It results that the geodesic distance between two points on the manifold is approximated by the shortest path that exists between these two points on the graph. We will use the Floyd-Warshall algorithm \cite{floyd}. The goal of this algorithm is to look for the shortest paths between all pairs of vertices in a weighted graph, where weights can be positive or negative. The complexity of this algorithm when computing the shortest paths between any 2 vertices is $O(N^{3})$, and $N$ is the number of vertices. In the context of image retrieval, an additional step is joined to the SM step, where the  Floyd–Warshall  algorithm  is  applied to the  matrix  of  distances (D) to compute  the  shortest  paths between all vertices (images). The input matrix (D) represents  the distance between the images existing in the dataset and is calculated using the KLD. It results that all paths that connect the vertices to each other in the weighted graph are initialized by their KLD measure. Then, the GD is approximated by the shortest path method.

\section{Experimental results}
\begin{figure}[htb!]
\begin{center}
\includegraphics[width=8cm]{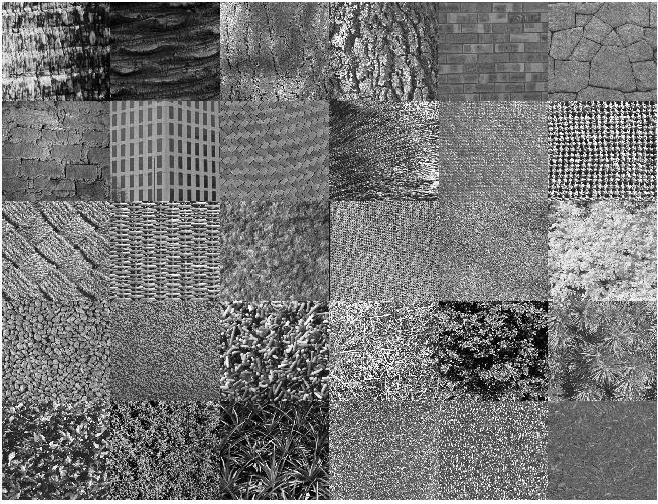}
\end{center}
\caption{30 texture images from the Vistex database.}
\label{VisTex}
\end{figure}

\begin{figure}[htb!]
\begin{center}
\includegraphics[width=8cm]{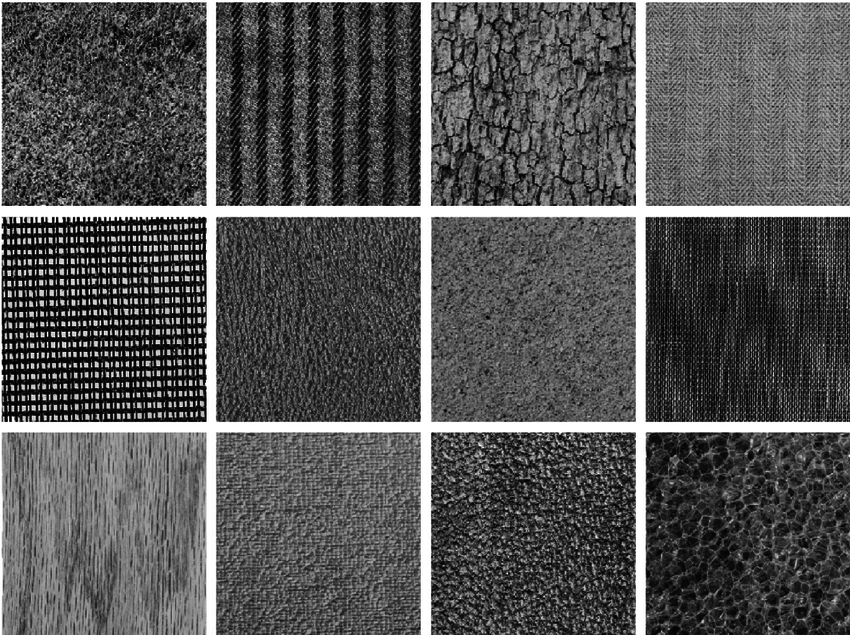}
\end{center}
\caption{12 texture images from the Brodatz database.}
\label{brodatz}
\end{figure}

To assess the performance of the proposed approach and its potential, we conduct series of experiments on two popular databases considering the image retrieval application:
\begin{itemize}
 \item \textbf{Dataset1}: The first dataset is a collection of 40 classes from the Vistex database \cite{vistex}. Each class contains 16 images of 512 x 512 pixels that results in a dataset of 640 images. (Figure. \ref{VisTex}). 
 \item \textbf{Dataset2}: The second dataset is the Brodatz \cite{brodatz} database which contains 111 gray-level texture images (Figure. \ref{brodatz}). Each of those images is divided into 16 of 640 x 640 pixel sub-images which results in a dataset of 1776 images.
\end{itemize}

The texture retrieval framework begins with the Feature extraction (FE) step or in other terms, the construction of the manifold. It starts by converting all the color images that exist in the dataset to grayscale images. Then each image  is  decomposed  into  frequency  subbands by the DTCWT wavelet decomposition \cite{mallat} using the Q-Shift (14,14) filters when the decomposition levels are greater or equal than two, and (13,19) near-orthogonal filters for level one. Afterward, we will rely on the Gamma and Weibull distributions to model the histogram of subbands coefficients and estimate their parameters. The estimated parameters for all subbands will construct the signature (point of the manifold) of a single image.\\
The similarity measurement step, where a distance (here the GD) is calculated between the points of the manifold, especially between signatures of two images $I_Q$ (query image) and $I_T$ (target image) as:

\begin{equation}\begin{split}
GD(I_{Q} \parallel I_{T}) =  \sum_{j=1}^{6k} GD ((\alpha_{j}^{Q}, \beta_{j}^{Q}, \lambda_{j}^{Q}, \Sigma_{j}^{Q})\\ \parallel (\alpha_{j}^{T}, \beta_{j}^{T}, \lambda_{j}^{T}, \Sigma_{j}^{Q}))
\end{split}.\end{equation}
The retrieval accuracy is estimated by presenting every point of the manifold as a query image, for which we retrieve the most similar images. The acquired retrieval rates for all images are averaged to compute the average retrieval rate (ARR) by this formula:
\begin{equation} ARR(K) = \frac{1}{N_{t}  N_{R}} \sum_{q=1}^{N_{t}} n_{q}(K)\bigl\vert_{K \geq N_{R}}. \end{equation} 
where $N_{t}$ and $N_{R}$ represent the total number of images in the dataset and the number of relevant images for each query. For each query image q, $n_{q}(K)$ is the number of correctly retrieved images among the K retrieved ones (i.e K best matches).
\\To measure the competitiveness of the proposed approaches Gamma+GDFloyd and Weibull+GDFloyd, we compare them  with other literature methods. In this regard, we will investigate the following approaches: 
\begin{itemize}

\item \textbf{Weibull+KLD} \cite{wblkld}: It represents the histogram of the wavelet coefficient subbands through the Weibull distribution combined with a closed-form of the KLD, 
 \item \textbf{Gamma+KLD} \cite{gammakld}:  It models the wavelet coefficient subbands using the Gamma distribution with closed-form of KLD as a similarity measure,
 \item  \textbf{GGD+KLD} \cite{dovett}: In this method the wavelet coefficient subbands are modeled using Generalized gaussian distributions and it uses a closed-form expression of the KLD during the SM step. 
 
 \item \textbf{Gamma+GDSKLD} \cite{zakariae}: In this method, the wavelet subbands are represented through the Gamma distribution and the GD is approximated by the Symmetric KLD,
 
 \item \textbf{Weibull+GDSKLD}: Same as Gamma+GDSKLD, wavelet subbands are modeled through the Weibull distribution, and the GD is computed approximated by the Symmetric KLD,

\end{itemize}

\begin{table*}[h]
\caption{ The Average Retrieval Rates (\%) concerning Dataset1 and Dataset2 .}
\label{vistexbrodatzARR}
\begin{center}
\renewcommand{\arraystretch}{1.5}

\begin{tabular}{c|c|c||c|c||c|c}
\hline
 &  \multicolumn{2}{c||}{1-Level}
 &  \multicolumn{2}{c||}{2-Level}	
 &  \multicolumn{2}{c}{3-Level} \\
\hline
 &Dataset1 & Dataset2 & Dataset1 & Dataset2 & Dataset1 & Dataset2 \\
\hline
Gamma+GDFloyd & 75.76 & 64.92 & 81.70 & 73.18 & \bf 84.62 & \bf 75.74 \\
\hline
Weibull+GDFloyd & 76.16 & 65.11 &  81.98 & 73.04 & \bf 85.16  & \bf 75.63  \\
\hline
Gamma+KLD\cite{gammakld} & 71.43 & 61.74 & 77.02 & 69.82 & 80.77 & 73.11 \\
\hline
Weibull+KLD \cite{wblkld} & 71.38 & 61.93 & 77.31 & 69.71 & 81.01 & 73.01 \\
\hline
GGD+KLD \cite{dovett} & 70.55 & 62.24 & 76.17 & 69.25 & 79.70 & 72.35 \\
\hline
 
\end{tabular}
\end{center}

\label{tableau}	
\end{table*} 
Table \ref{vistexbrodatzARR} shows the ARRs of the proposed methods in comparison with the literature methods considering Dataset1 and Dataset2. It is to remark that the proposed approaches lead to higher performances. It means that taking into consideration the extra information given by the geometrical properties of the statistical manifolds improves the retrieval rate. It also confirms the effectiveness of the graph-based approximation of the GD. This is mostly a result of the good representation of the manifold by the graph approach. this latter preserves its structure and geometrical properties. \\The performance of our approach is also explained by the use of the GD, which is a real distance compared to the KLD which is not symmetric and does not satisfy the triangular inequality. \\Moreover, it is to note that more levels of DTCWT decomposition improves the performance of the ARR. Be that as it may, the underlying two decomposition levels represent by far most of the representation power since the higher improvement is accomplished from 1-Level to 2-Level. The third decomposition level gets little improvement when added together with the underlying two levels. Moving towards another level is not advantageous since it won't improve results. \\The difference in performances of all methods considering the two datasets is explained by the number of images (640 vs 1776), and the heterogeneous nature of the images that makes the retrieval harder in case of Dataset2.
\begin{table}
\begin{center}
\renewcommand{\arraystretch}{1.8}
\caption{Average Retrieval Rate (\%) using different approximation methods of the Geodesic Distance for three DTCWT scales, on the Dataset1 and Dataset2}
\label{GDapprox}
\begin{tabular}{p{3cm}|c|c}
\hline

 &  Dataset1 & Dataset2\\
\hline
 Gamma+GDFloyd & \bf  84.62 & \bf 75.74\\
\hline
 Weibull+GDFloyd & \bf 85.16 & \bf 75.63\\
\hline
 Gamma+GDSKLD\cite{zakariae} & 80.94 & 73.12\\
\hline
 Weibull+GDSKLD\cite{zakariae} & 81.64 &    72.60
 \\
\hline

\end{tabular}
\end{center}
\end{table}
\\Table \ref{GDapprox} compares the proposed  method to another GD approximation. In the latter case, the GD is approximated by the square root of the double of the Symmetric KLD (SKLD), since it is proved for distributions that lie infinitesimally close on the probabilistic manifold \cite{kullback}. We note that the graph-based approach outperforms the SKLD based approximation. This proofs the already made conclusion on Table \ref{vistexbrodatzARR}, and supports that the graph representation conserves the geometrical properties of the manifold. 
\begin{figure}[h]
\begin{center}
\includegraphics[width=8cm]{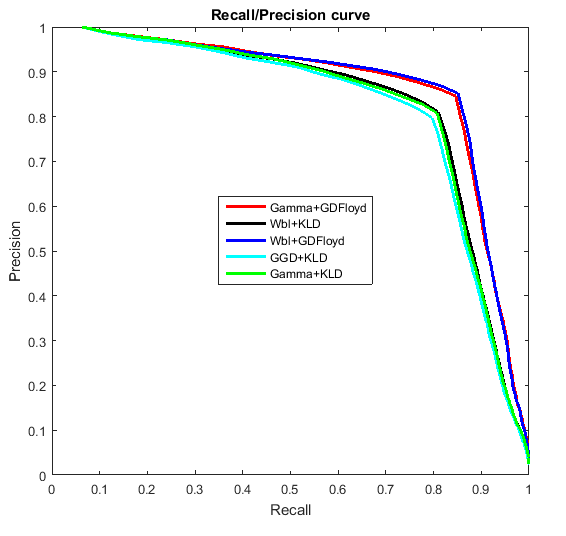}
\end{center}
\caption{Recall/Precision curves obtained on Dataset1.}
\label{Curve1}
\end{figure}
\\Figure. \ref{Curve1} shows the Recall/Precision curves obtained using all approaches on Dataset1, considering the third level of decomposition of DTCWT. We note that the proposed approaches (Gamma+GDFloyd and Weibull+GDFloyd) are clearly outperforming the other literature methods all along the curve. As aforesaid, this is due to the extra information given by the geometrical properties of the statistical manifolds and exploited through the GD, which improves the retrieval rate. 

\begin{figure}[h]
\begin{center}
\includegraphics[width=8cm]{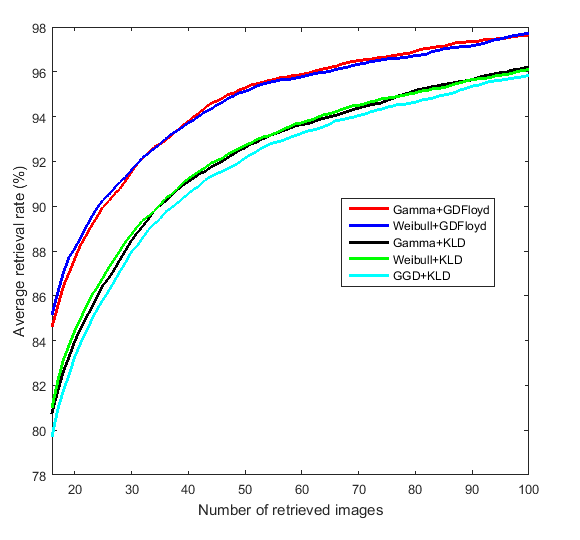}
\end{center}
\caption{Retrieval effectiveness with respect to the number of top N matches considered on theDataset1}
\label{Curve2}
\end{figure}

Figure. \ref{Curve2} shows the retrieval precision as a function of the number of retrieved images for Dataset1, considering the third level of decomposition of DTCWT. We notice that when considering 16 retrieved images, all the methods gives the same performance as presented in Table \ref{vistexbrodatzARR}. Nevertheless, the proposed approach shows better performance from the beginning and stays in parallel with the other literature methods all along the curve as the number becomes near to 100 retrieved images. This gives additional proof of the effectiveness of the proposed approaches and show the importance of using the geometrical properties of the statistical manifolds and the accuracy of the graph-based method in exploiting these properties.

\section{Conclusion}

In  this  work,  we  proposed the intrinsic geodesic distance as an accurate similarity measurement instead of the statistical Kullback-Leibler divergence.  Our  main goal was to  exploit  the properties of the statistical Gamma and Weibull manifolds by using the GD through a graph-based method that preserves most of the geometrical properties. We  succeeded  to propose a graph-based method to approximate the GD in the case of Gamma and Weibull manifolds, that outperforms the previous GD approximations. The experimental results indicate that the GD achieves higher performances than the Kullback-Leibler divergence when using the graph-based  approximation.  Moreover, the  advantage with the GD  remains important  as it fulfills the conditions of a real distance and permits to make geometric interpretations on the studied manifold. Our future works will be devoted to the geometric studies of other manifolds with an application to color texture retrieval.

%

\bibliographystyle{unsrt}
\bibliography{biblio}

\end{document}